\documentclass{article}

\usepackage[nonatbib, final]{neurips}
\usepackage[numbers]{natbib}

\makeatletter
\renewcommand{\@noticestring}{\centering}
\makeatother

\usepackage[export]{adjustbox}
\usepackage[inline, shortlabels]{enumitem}
\usepackage[T1]{fontenc}
\usepackage{hyperref}
\usepackage{microtype}
\usepackage{pifont}
\usepackage{xcolor}
\usepackage{xurl}
\usepackage{graphicx}
\usepackage{booktabs}
\usepackage{tabularray}
\usepackage{listings}
\usepackage{amsmath, amsfonts}
\usepackage{amsthm}
\usepackage{nicefrac}

\theoremstyle{definition}
\newtheorem{definition}{Definition}

\UseTblrLibrary{booktabs}

\makeatletter
\newcommand{\ssymbol}[1]{\@fnsymbol{#1}}
\newcommand{\romanNumeral}[1]{\expandafter\@slowromancap\romannumeral #1@}
\makeatother

\usepackage{subfig}
\usepackage{multirow}
\usepackage{algorithm}
\usepackage{algorithmic}

\graphicspath{{figures/}}

\newcommand{\acar}{\textsc{ACAR}}
\newcommand{\teamllm}{\textsc{TeamLLM}}

\title{ACAR: Adaptive Complexity Routing for Multi-Model Ensembles\\with Auditable Decision Traces}

\author{
    Ramchand Kumaresan
}

\begin{document}
\raggedbottom

\maketitle

%==============================================================================
% ABSTRACT
%==============================================================================
\begin{abstract}
We present \acar{} (Adaptive Complexity \& Attribution Routing) as a measurement framework for studying multi-model orchestration under auditable conditions.
\acar{} uses self-consistency variance ($\sigma$) computed from N=3 probe samples to route tasks across single-model, two-model, and three-model execution modes, implemented atop \teamllm{}, a deterministic substrate with immutable artifacts and complete decision traces.
We evaluate across 1,510 tasks spanning four benchmarks (MathArena, Reasoning Gym, LiveCodeBench, SuperGPQA) with Claude Sonnet 4, GPT-4o, and Gemini 2.0 Flash, producing 7,550+ auditable runs.
\textbf{What holds}: $\sigma$-based routing achieves 55.6\% accuracy, exceeding the two-model baseline (54.4\%) while avoiding full ensembling on 54.2\% of tasks; the mechanism is model-agnostic and requires no learned components.
\textbf{What does not hold}: (1) Retrieval augmentation \emph{decreased} accuracy by 3.4 percentage points---median retrieval similarity was only 0.167, demonstrating that experience injection without semantic alignment introduces harmful noise rather than grounding.
(2) When models agree on incorrect answers ($\sigma$=0), no downstream ensemble can recover; this ``agreement-but-wrong'' failure mode is intrinsic to self-consistency and bounds achievable accuracy at 8pp below full ensembling.
(3) Attribution estimates based on proxy signals (response similarity, entropy) showed weak correlation with ground-truth leave-one-out values; practical attribution requires explicit counterfactual computation.
This paper documents what assumptions fail in practice, providing falsifiable baselines for future work on routing, retrieval, and multi-model attribution.
\end{abstract}

%==============================================================================
% 1. INTRODUCTION
%==============================================================================
\section{Introduction}

\subsection{Background and Motivation}

Large language models (LLMs) have achieved strong performance across diverse tasks, but deploying them effectively requires navigating a fundamental trade-off between quality and cost. A single model may produce incorrect or incomplete answers, while running multiple models in parallel (ensembling) improves reliability but multiplies inference costs. This tension---\emph{how to allocate compute before knowing whether a task is easy or hard}---is central to practical LLM deployment.

\paragraph{Why the problem matters.}
Organizations deploying LLMs face heterogeneous workloads where task difficulty varies substantially. Simple factual queries may require only one model, while complex reasoning tasks benefit from multiple perspectives. Naive strategies---always using one model (cheap but unreliable) or always using many (reliable but expensive)---leave significant value unrealized. A principled routing mechanism that allocates compute adaptively could reduce costs on easy tasks while preserving quality on hard ones.

\paragraph{Why existing solutions are insufficient.}
Current approaches to multi-model orchestration fall into two categories, each with limitations:

\begin{enumerate}[leftmargin=*, itemsep=2pt]
    \item \textbf{Learned routers}~\citep{routerbench, frugalgpt, routellm} train classifiers to predict which model will succeed on a given query. These achieve high routing accuracy but introduce distribution shift between training and deployment, lack interpretable decision traces, and do not address credit assignment when multiple models contribute.

    \item \textbf{Observability platforms} provide post-hoc dashboards showing model performance but do not influence routing decisions at execution time. They enable debugging but not adaptive resource allocation.
\end{enumerate}

Neither approach provides the combination of (a) cost-aware routing, (b) measurable attribution when models collaborate, and (c) deterministic reproducibility required for rigorous experimentation.

\paragraph{Why the problem is challenging.}
Three factors make principled multi-model routing difficult:

\begin{itemize}[leftmargin=*, itemsep=2pt]
    \item \textbf{Task difficulty is latent}: We cannot directly observe whether a task is ``hard'' before attempting it. Difficulty must be estimated from proxy signals.
    \item \textbf{Output equivalence is non-trivial}: Determining whether two model outputs are semantically equivalent (especially for code) requires domain-specific comparison logic.
    \item \textbf{Attribution requires counterfactuals}: Assigning credit to individual models in an ensemble requires knowing what would have happened without each model---expensive counterfactual runs.
\end{itemize}

\subsection{Approach and Contributions}

We propose \acar{} (Adaptive Complexity \& Attribution Routing), which uses \emph{self-consistency variance} ($\sigma$) as a task difficulty signal. The intuition is simple: when multiple samples from a fast model agree on an answer, the task is likely easy and a single model suffices; when samples disagree, the task is ambiguous and benefits from diverse model perspectives.

We deliberately use a heuristic $\sigma$ rather than a learned router. This design choice sacrifices some routing accuracy for three properties we consider essential for credible measurement: (1) no distribution shift between training and deployment, (2) fully interpretable decision traces, and (3) immunity to benchmark-specific overfitting.

\paragraph{Contributions.}
\begin{itemize}[leftmargin=*, itemsep=2pt, topsep=0pt]
    \item We introduce \acar{}, an adaptive routing mechanism using $\sigma$-based task difficulty estimation that achieves 55.6\% accuracy while avoiding full ensembling on 54.2\% of tasks.
    \item We demonstrate that retrieval augmentation with low-quality stores \emph{hurts} performance (-3.4pp), providing actionable negative results: similarity thresholds $>$0.7 are required.
    \item We release \teamllm{}, a deterministic execution substrate with 7,550+ auditable runs, enabling reproducible multi-model research.\footnote{Code and artifacts are publicly available at \url{https://github.com/mechramc/ACAR-TeamLLM}.} All figures in this paper can be regenerated from released artifacts.
\end{itemize}

%==============================================================================
% 2. RELATED WORK
%==============================================================================
\section{Related Work}

We position this work at the intersection of three research areas: multi-model routing, cost-aware inference, and reproducible benchmarking. We summarize each area and clarify how \acar{} differs.

\subsection{Multi-Model Routing and Orchestration}

\paragraph{Learned routers.}
RouterBench~\citep{routerbench} introduced a benchmark for evaluating LLM routing systems, comparing learned classifiers that predict which model will succeed on a given query. FrugalGPT~\citep{frugalgpt} proposed cascading strategies that route to cheaper models first and escalate only on failure. RouteLLM~\citep{routellm} trains routers using preference data from human evaluations.

These systems optimize for routing accuracy---maximizing the probability of selecting the best model. \acar{} differs in three ways: (1) we use self-consistency rather than learned classifiers, avoiding distribution shift; (2) we log complete decision traces for auditability; (3) we explicitly measure and report failure modes rather than optimizing a single metric.

\paragraph{Multi-agent systems.}
ReAct~\citep{react} and related work coordinate tool-augmented agents but focus on single-model reasoning augmented with external tools, not multi-model ensembles. Mixture-of-Experts~\citep{moe} routes at the token level within a single model architecture, orthogonal to our inter-model routing.

\subsection{Cost-Aware and Adaptive Inference}

Several works address the cost-quality trade-off in LLM inference. Speculative decoding uses small models to draft tokens that larger models verify. Early-exit strategies terminate computation when confidence is high. These operate within a single model or model pair; \acar{} addresses routing across heterogeneous models from different providers.

The key distinction is that \acar{} treats cost awareness as a \emph{measurement} problem: we seek to understand when adaptive routing helps and when it fails, not to maximize a cost-quality metric.

\subsection{Reproducible Benchmarking Frameworks}

Benchmark contamination and evaluation inconsistency have motivated calls for more rigorous evaluation infrastructure. LiveCodeBench provides execution-verified code tasks with temporal splits. SuperGPQA tests broad knowledge with multiple-choice questions.

\acar{} contributes to this area by providing: (1) deterministic execution with logged seeds and environment fingerprints, (2) immutable artifacts enabling independent verification, and (3) explicit reporting of negative results alongside positive findings.

\paragraph{How this work differs.}
Unlike learned routing systems, \acar{} prioritizes auditability and reproducibility over routing accuracy. Unlike observability platforms, \acar{} makes routing decisions at execution time based on measurable signals. Unlike benchmark papers, we report both what works and what fails, with complete artifacts for reproduction.

%==============================================================================
% 3. METHOD
%==============================================================================
\section{Method}

\subsection{\teamllm{} Execution Substrate}

\teamllm{} is a research-grade execution substrate designed for auditable multi-model experiments. It enforces three invariants:

\begin{enumerate}[leftmargin=*, itemsep=2pt]
    \item \textbf{Deterministic execution}: Every run captures: random seed, prompt template hash, rubric version, model identifiers, and environment fingerprint. Re-execution with identical inputs produces identical outputs.
    \item \textbf{Immutable artifacts}: All responses, evaluations, and decision traces are append-only. Modifications create new versioned records; existing records cannot be altered.
    \item \textbf{Forward-only state machine}: Run status progresses through defined states (PENDING $\to$ EXECUTING $\to$ VERIFYING $\to$ COMPLETED) with no rollback transitions.
\end{enumerate}

Artifacts are stored as structured JSONL files (\texttt{runs.jsonl}) containing per-task decision traces. All figures in this paper can be regenerated from these artifacts using provided scripts.

\subsection{\acar{}: Adaptive Complexity Routing}

\acar{} routes tasks to execution modes based on \emph{self-consistency variance} ($\sigma$), a measure of agreement among multiple samples from a fast ``probe'' model.

\subsubsection{Definitions}

Let $\mathcal{T}$ be a task (prompt + expected output format). Let $M_{\text{probe}}$ be a fast model used for difficulty estimation (we use Gemini 2.0 Flash). Let $\textsc{Extract}(r)$ denote an answer extraction function that maps a model response $r$ to a canonical answer representation.

\begin{definition}[Self-Consistency Variance]
Given $N=3$ independent samples $r_1, r_2, r_3$ from $M_{\text{probe}}$ on task $\mathcal{T}$, let $a_i = \textsc{Extract}(r_i)$ be the extracted answers. The self-consistency variance is:
\begin{equation}
\sigma = \frac{|\{a_1, a_2, a_3\}| - 1}{2}
\end{equation}
where $|\{a_1, a_2, a_3\}|$ counts distinct answers. $\sigma \in \{0.0, 0.5, 1.0\}$.
\end{definition}

\begin{definition}[Execution Mode]
The execution mode $\mathcal{M}(\sigma)$ maps variance to model count:
\begin{align}
\mathcal{M}(\sigma) = \begin{cases}
\texttt{single\_agent} & \text{if } \sigma = 0.0 \text{ (all agree)} \\
\texttt{arena\_lite} & \text{if } \sigma = 0.5 \text{ (2/3 agree)} \\
\texttt{full\_arena} & \text{if } \sigma = 1.0 \text{ (all differ)}
\end{cases}
\end{align}
\end{definition}

\subsubsection{Algorithm}

Algorithm~\ref{alg:acar} presents the complete \acar{} routing procedure.

\begin{algorithm}[t]
\caption{\acar{} Routing Procedure}
\label{alg:acar}
\begin{algorithmic}[1]
\REQUIRE Task $\mathcal{T}$, probe model $M_{\text{probe}}$, ensemble models $\{M_1, M_2, M_3\}$
\ENSURE Final answer $a^*$, decision trace $D$

\STATE \textbf{Phase 1: Difficulty Estimation}
\FOR{$i = 1$ to $3$}
    \STATE $r_i \gets M_{\text{probe}}(\mathcal{T})$ \COMMENT{Sample from probe model}
    \STATE $a_i \gets \textsc{Extract}(r_i)$ \COMMENT{Extract canonical answer}
\ENDFOR
\STATE $\sigma \gets (|\{a_1, a_2, a_3\}| - 1) / 2$ \COMMENT{Compute variance}

\STATE \textbf{Phase 2: Adaptive Routing}
\IF{$\sigma = 0.0$}
    \STATE $a^* \gets a_1$ \COMMENT{All agree: use single answer}
    \STATE $\text{mode} \gets \texttt{single\_agent}$
\ELSIF{$\sigma = 0.5$}
    \STATE $a^* \gets \textsc{MajorityVote}(\{a_1, a_2, a_3\})$
    \STATE Execute $M_1, M_2$ for verification
    \STATE $\text{mode} \gets \texttt{arena\_lite}$
\ELSE
    \STATE Execute all models $\{M_1, M_2, M_3\}$
    \STATE $a^* \gets \textsc{JudgeSelect}(\{r_{M_1}, r_{M_2}, r_{M_3}\})$
    \STATE $\text{mode} \gets \texttt{full\_arena}$
\ENDIF

\STATE \textbf{Phase 3: Logging}
\STATE $D \gets \{\mathcal{T}, \sigma, \text{mode}, a^*, \text{timestamp}, \text{cost}\}$
\STATE Append $D$ to immutable trace

\RETURN $a^*$, $D$
\end{algorithmic}
\end{algorithm}

\subsubsection{Design Rationale}

Several design choices merit explanation:

\paragraph{Why $N=3$ samples?} Three samples provide the minimal information to distinguish consensus (all agree), partial agreement (2/3 agree), and disagreement (all differ). Larger $N$ would provide finer granularity but increase probe cost.

\paragraph{Why discrete $\sigma$ rather than continuous?} A discrete routing signal ensures deterministic, interpretable decisions. Continuous signals would require threshold tuning, introducing hyperparameters that could overfit to specific benchmarks.

\paragraph{Why not a learned router?} Learned routers achieve higher routing accuracy but introduce distribution shift and lack interpretability. For a \emph{measurement} framework, we prioritize auditability over optimization.

\subsubsection{Configurations}

We evaluate two \acar{} variants:

\paragraph{\acar{}-U (Ult only)} uses $\sigma$-based routing without retrieval augmentation.

\paragraph{\acar{}-UJ (Ult + Jungler)} adds asynchronous retrieval from an experience store, injecting similar past examples into prompts before model dispatch. The Jungler component retrieves experiences with similarity above a threshold (we use 0.0, i.e., any match). As we show in Section~\ref{sec:negative}, this design choice leads to negative results.

\begin{figure}[t]
\centering
\includegraphics[width=0.85\columnwidth]{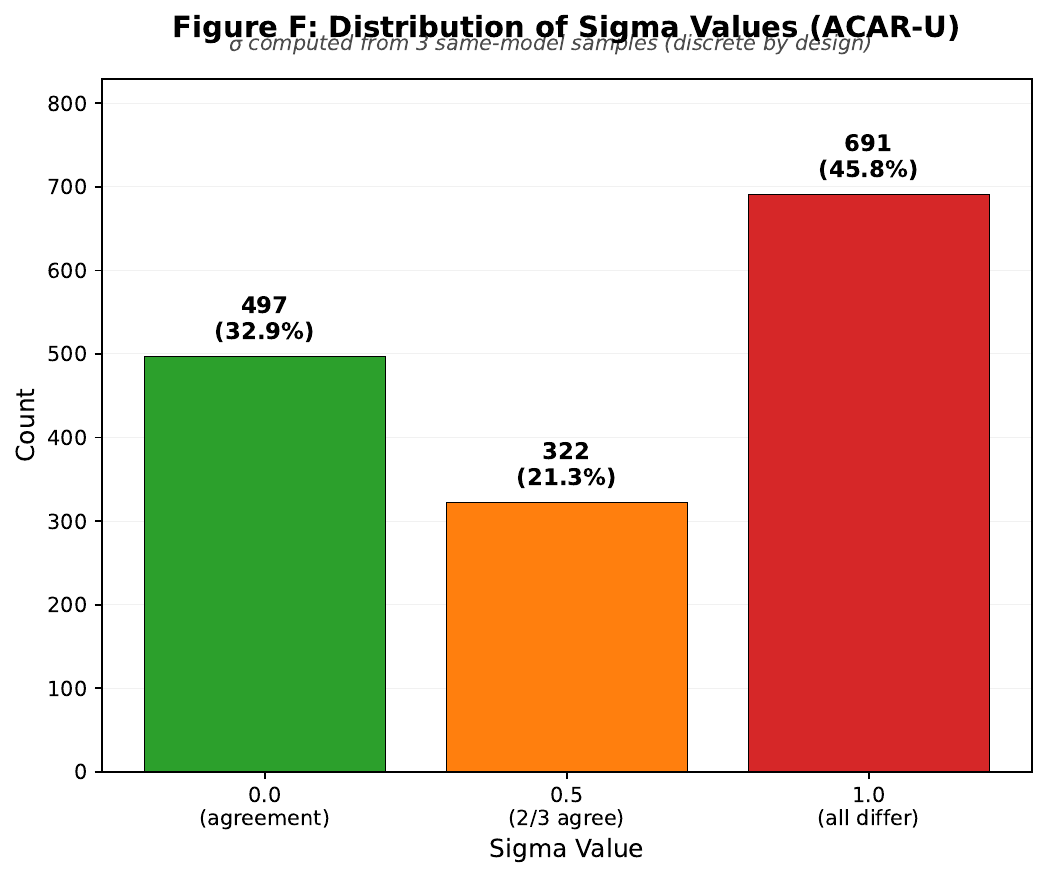}
\caption{\textbf{Distribution of $\sigma$ across 1,510 tasks.} Task difficulty is bimodal: 32.9\% show full agreement ($\sigma$=0.0), while 45.8\% show complete disagreement ($\sigma$=1.0). This bimodality enables effective routing---easy tasks avoid expensive ensembling.}
\label{fig:sigma}
\end{figure}

%==============================================================================
% 4. EXPERIMENTAL SETUP
%==============================================================================
\section{Experimental Setup}

\subsection{Benchmarks}

We evaluate on 1,510 tasks spanning four benchmarks selected to cover diverse task types:

\begin{itemize}[leftmargin=*, itemsep=2pt]
    \item \textbf{MathArena} (60 tasks): Competition mathematics problems requiring multi-step reasoning. Highest difficulty in our suite.
    \item \textbf{Reasoning Gym} (250 tasks): Procedural reasoning tasks testing logical inference chains.
    \item \textbf{LiveCodeBench} (200 tasks): Code generation with execution-based verification. Answers are correct only if generated code passes test cases.
    \item \textbf{SuperGPQA} (1,000 tasks): Multiple-choice knowledge questions spanning diverse domains.
\end{itemize}

\subsection{Models}

All experiments use three frontier models at temperature 0 for deterministic evaluation:
\begin{itemize}[leftmargin=*, itemsep=1pt]
    \item Claude Sonnet 4 (Anthropic)
    \item GPT-4o (OpenAI)
    \item Gemini 2.0 Flash (Google) --- also used as probe model $M_{\text{probe}}$
\end{itemize}

\subsection{Configurations}

We compare five configurations:
\begin{itemize}[leftmargin=*, itemsep=1pt]
    \item \textbf{Single-Model}: Best single model only (Claude Sonnet 4)
    \item \textbf{Arena-2}: Two-model ensemble (Claude + GPT-4o)
    \item \textbf{Arena-3}: Three-model ensemble (all models, all tasks)
    \item \textbf{\acar{}-U}: Adaptive routing without retrieval
    \item \textbf{\acar{}-UJ}: Adaptive routing with retrieval augmentation
\end{itemize}

\subsection{Metrics}

\begin{itemize}[leftmargin=*, itemsep=1pt]
    \item \textbf{Accuracy}: Fraction of tasks with correct final answer
    \item \textbf{Cost}: Total API cost in USD
    \item \textbf{Escalation rate}: Fraction of tasks routed to each mode
\end{itemize}

%==============================================================================
% 5. RESULTS
%==============================================================================
\section{Results}

\subsection{Main Results}

\begin{table}[t]
\centering
\caption{\textbf{Overall accuracy on 1,510 tasks.} \acar{}-U exceeds Arena-2 while costing less. Arena-3 represents the quality ceiling. Arena-2 and Arena-3 show identical cost due to coordination overhead dominating marginal per-model costs.}
\begin{tabular}{@{}lrrr@{}}
\toprule
Configuration & Accuracy & Correct & Cost \\
\midrule
Single-Model & 45.4\% & 686/1510 & \$17.04 \\
Arena-2 & 54.4\% & 822/1510 & \$20.64 \\
\textbf{\acar{}-U} & \textbf{55.6\%} & \textbf{839/1510} & \textbf{\$20.34} \\
Arena-3 & 63.6\% & 961/1510 & \$20.64 \\
\bottomrule
\end{tabular}
\label{tab:main}
\end{table}

\begin{figure}[t]
\centering
\includegraphics[width=0.9\columnwidth]{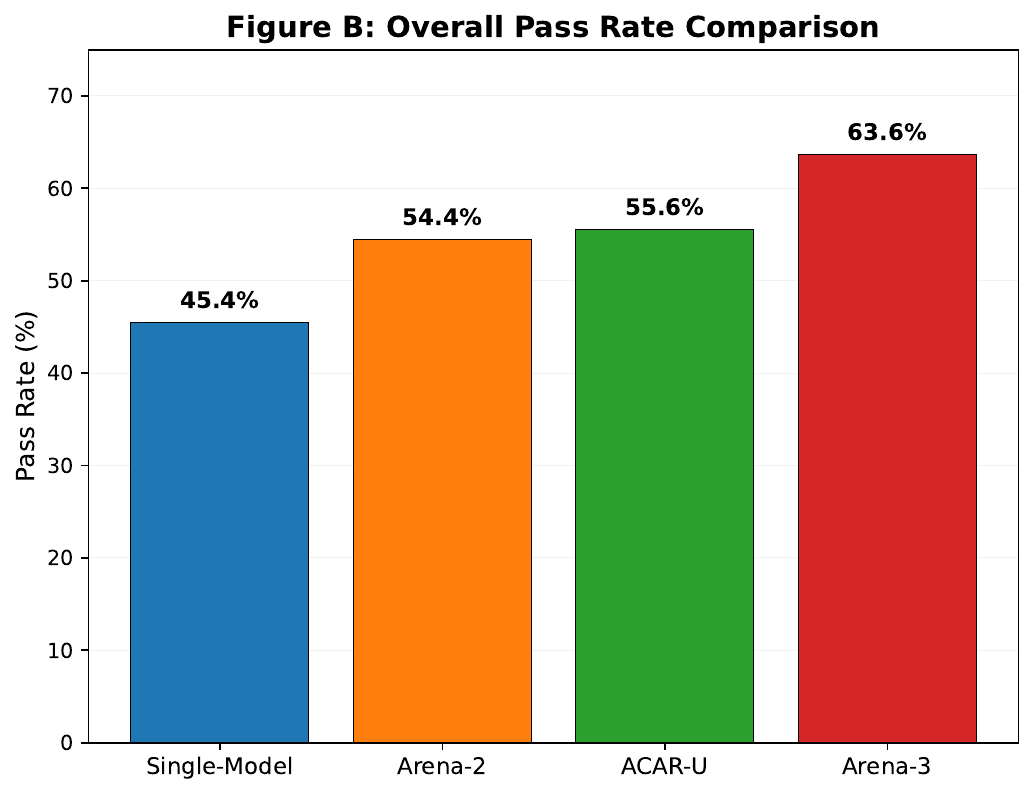}
\caption{\textbf{Overall pass rate comparison.} \acar{}-U (55.6\%) exceeds Arena-2 (54.4\%) with adaptive compute allocation. Arena-3 (63.6\%) represents the quality ceiling.}
\label{fig:overall}
\end{figure}

Table~\ref{tab:main} and Figure~\ref{fig:overall} show main results. \acar{}-U exceeds Arena-2 by 1.2 percentage points while costing 1.5\% less. The 8.0pp gap to Arena-3 reflects a fundamental limitation discussed in Section~\ref{sec:negative}.

\begin{figure}[t]
\centering
\includegraphics[width=0.9\columnwidth]{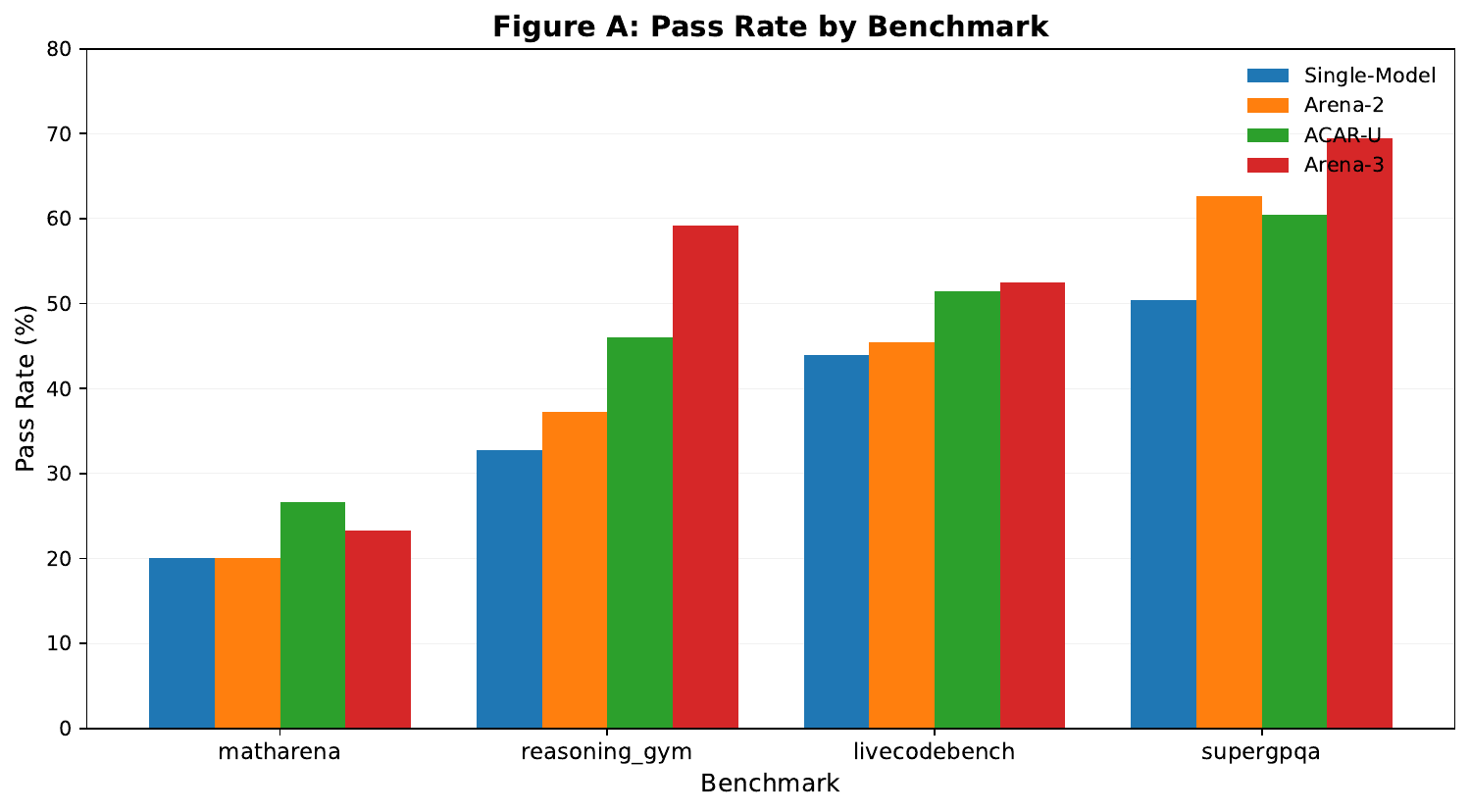}
\caption{\textbf{Pass rate by benchmark.} Performance varies by domain: SuperGPQA shows highest accuracy (60.5\%), while MathArena remains challenging (26.7\%).}
\label{fig:benchmark}
\end{figure}

Figure~\ref{fig:benchmark} shows per-benchmark results. \acar{}-U matches or exceeds Arena-2 on all benchmarks.

\subsection{Cost-Accuracy Trade-off}

\begin{figure}[t]
\centering
\includegraphics[width=0.9\columnwidth]{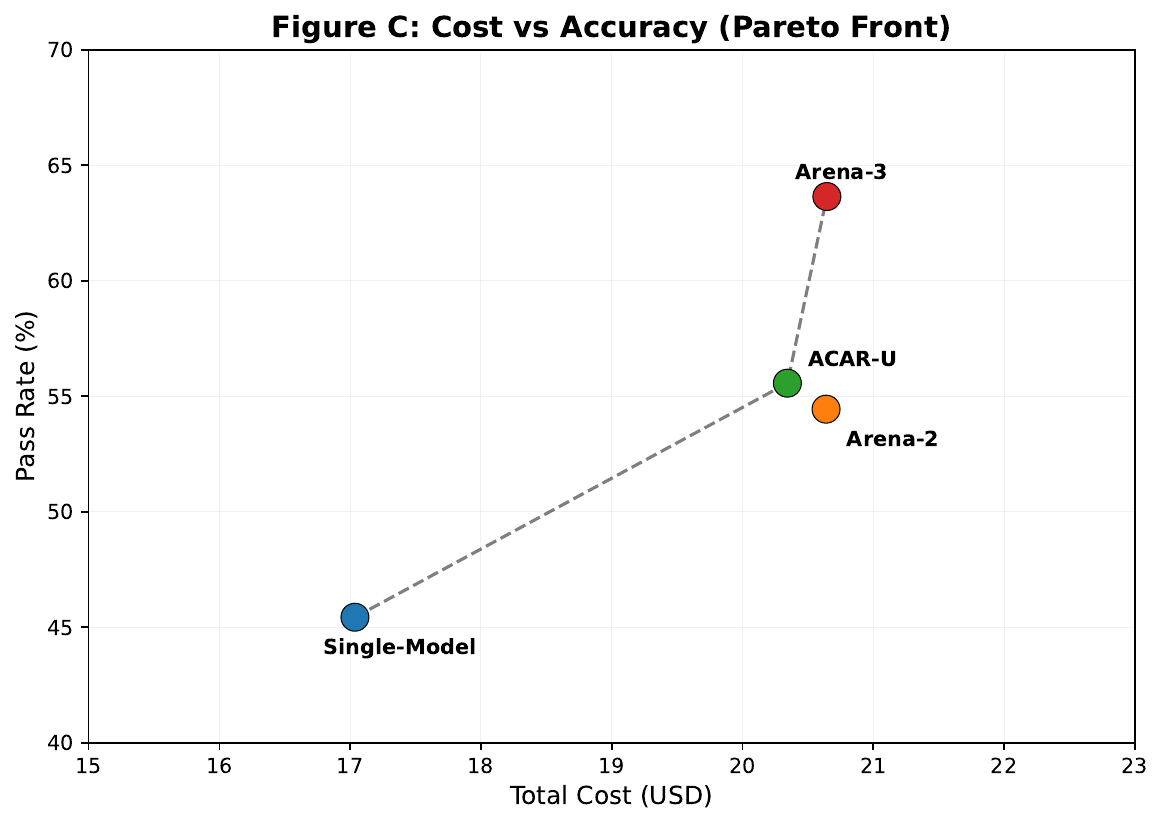}
\caption{\textbf{Cost vs.\ accuracy Pareto frontier.} \acar{}-U achieves better accuracy than Arena-2 at lower cost.}
\label{fig:pareto}
\end{figure}

Figure~\ref{fig:pareto} shows the cost-accuracy trade-off. \acar{}-U achieves a favorable position on the Pareto frontier.

\subsection{Escalation Behavior}

\begin{figure}[t]
\centering
\includegraphics[width=0.9\columnwidth]{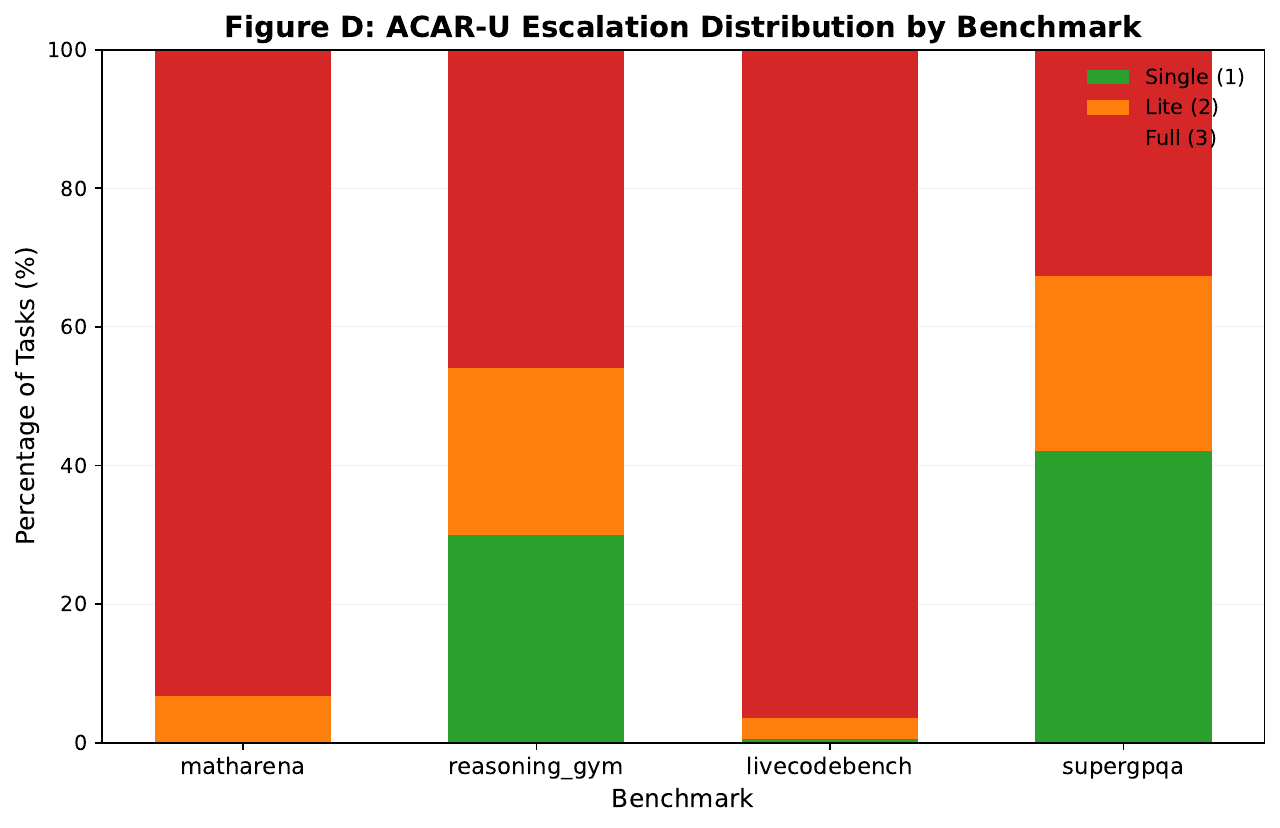}
\caption{\textbf{Escalation distribution by benchmark.} $\sigma$-routing adapts to task difficulty: SuperGPQA routes 42\% to single-agent, while MathArena (93\%) and LiveCodeBench (96\%) escalate to full-arena.}
\label{fig:escalation}
\end{figure}

\acar{}-U routes 32.9\% of tasks to single-agent, 21.3\% to arena-lite, and 45.8\% to full-arena. Figure~\ref{fig:escalation} shows this varies by benchmark: SuperGPQA (easiest) routes 42\% to single-agent, while LiveCodeBench escalates 96\% to full-arena due to non-canonical code outputs.

\begin{figure}[t]
\centering
\includegraphics[width=0.9\columnwidth]{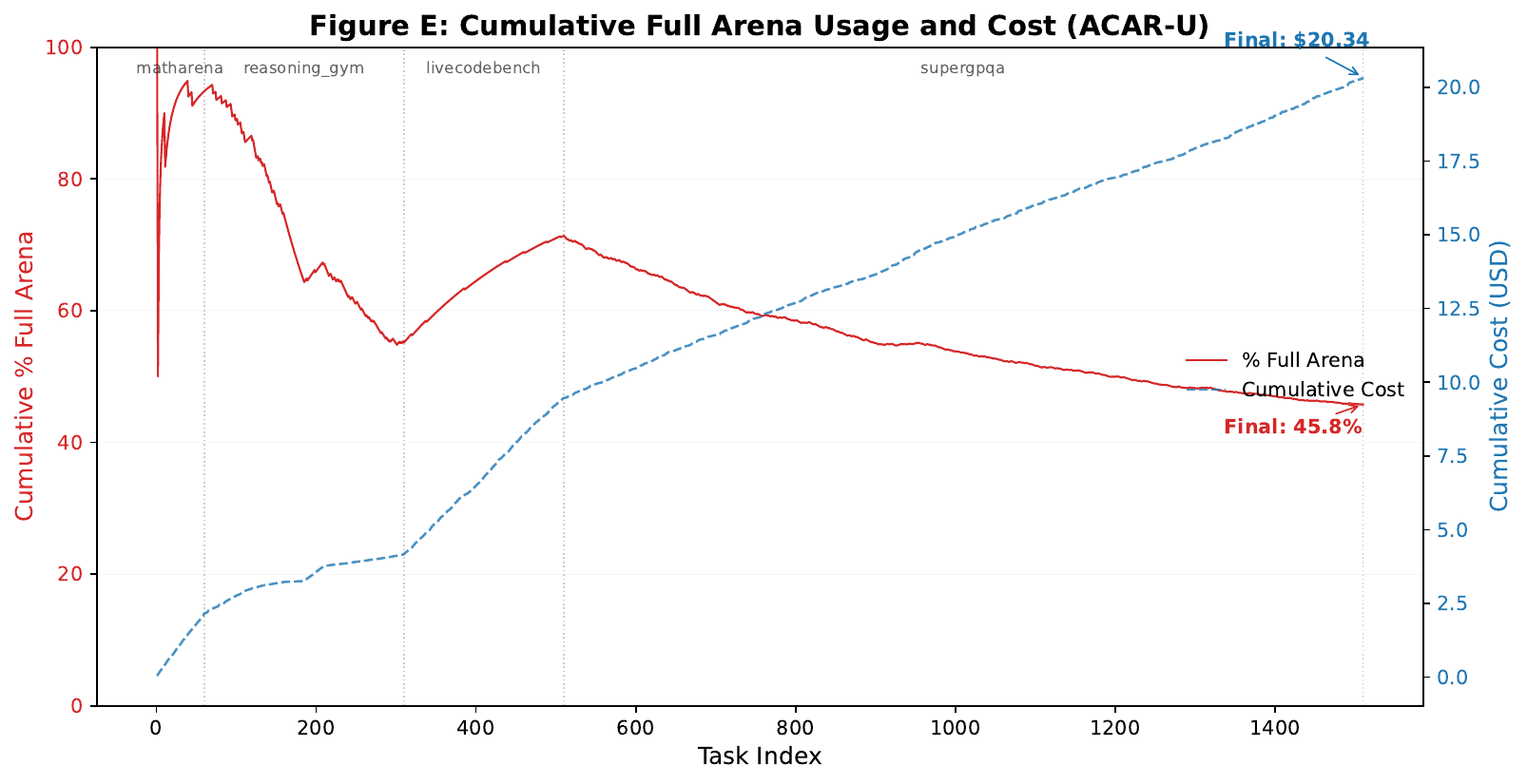}
\caption{\textbf{Cumulative full-arena usage.} \acar{}-U avoids full ensembling on 54.2\% of tasks.}
\label{fig:cumulative}
\end{figure}

Figure~\ref{fig:cumulative} shows cumulative full-arena usage. \acar{}-U avoids expensive full ensembling on the majority of tasks.

\subsection{Latency Analysis}

\begin{figure}[t]
\centering
\includegraphics[width=0.9\columnwidth]{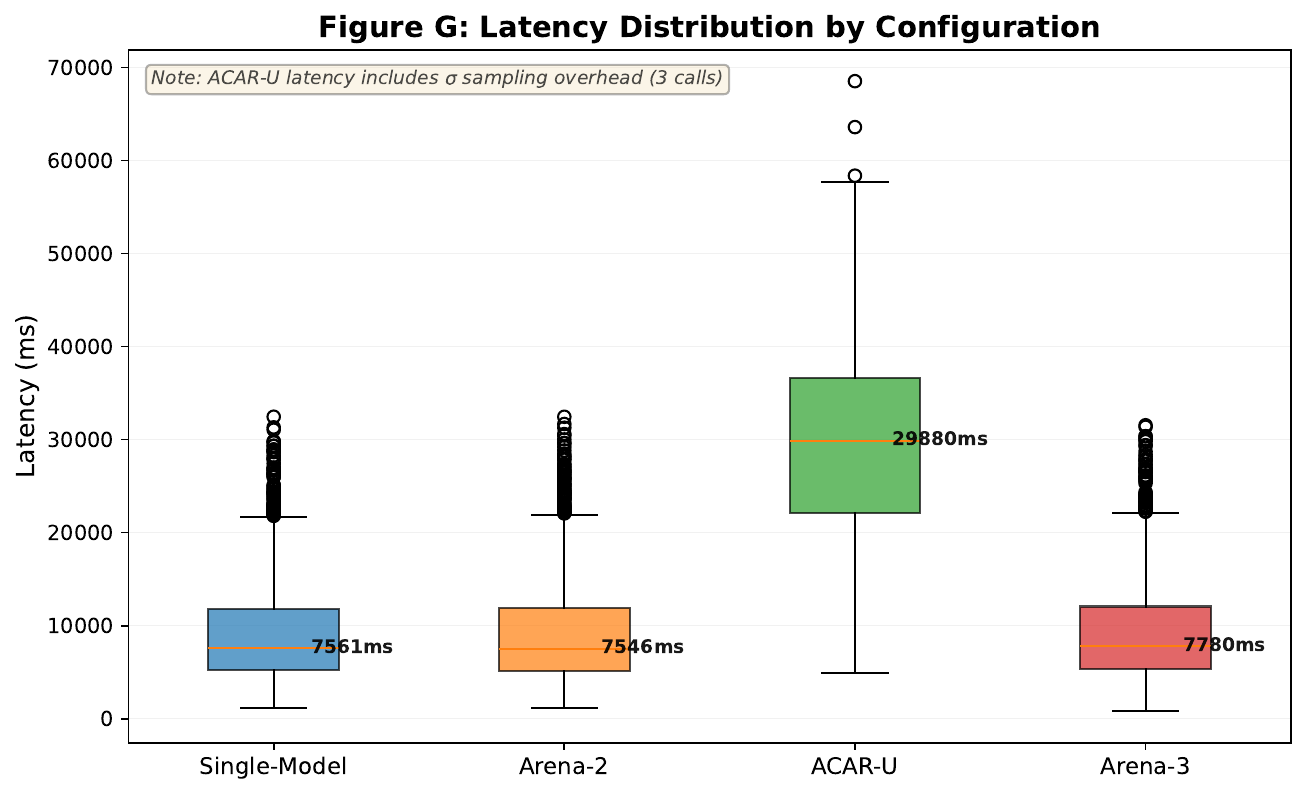}
\caption{\textbf{Latency distribution by configuration.} \acar{}-U achieves intermediate latency by routing easy tasks to faster single-model execution.}
\label{fig:latency}
\end{figure}

Figure~\ref{fig:latency} compares latency. Single-model execution provides lowest latency; full ensembling incurs coordination overhead. \acar{}-U achieves intermediate latency.

%==============================================================================
% 6. NEGATIVE RESULTS
%==============================================================================
\section{Negative Results and Failure Modes}
\label{sec:negative}

We report three significant negative findings. These are not incidental observations but systematic failures that bound the applicability of our approach.

\subsection{Retrieval Augmentation Hurts Performance}

\begin{table}[h]
\centering
\small
\caption{\textbf{\acar{}-UJ vs \acar{}-U.} Retrieval augmentation hurts across all benchmarks.}
\begin{tabular}{@{}lrrr@{}}
\toprule
Benchmark & \acar{}-U & \acar{}-UJ & $\Delta$ \\
\midrule
MathArena & 26.7\% & 21.7\% & -5.0pp \\
Reasoning Gym & 46.0\% & 44.0\% & -2.0pp \\
LiveCodeBench & 51.5\% & 47.5\% & -4.0pp \\
SuperGPQA & 60.5\% & 57.3\% & -3.2pp \\
\midrule
\textbf{Overall} & \textbf{55.6\%} & \textbf{52.4\%} & \textbf{-3.4pp} \\
\bottomrule
\end{tabular}
\label{tab:uj}
\end{table}

\begin{figure}[t]
\centering
\includegraphics[width=0.9\columnwidth]{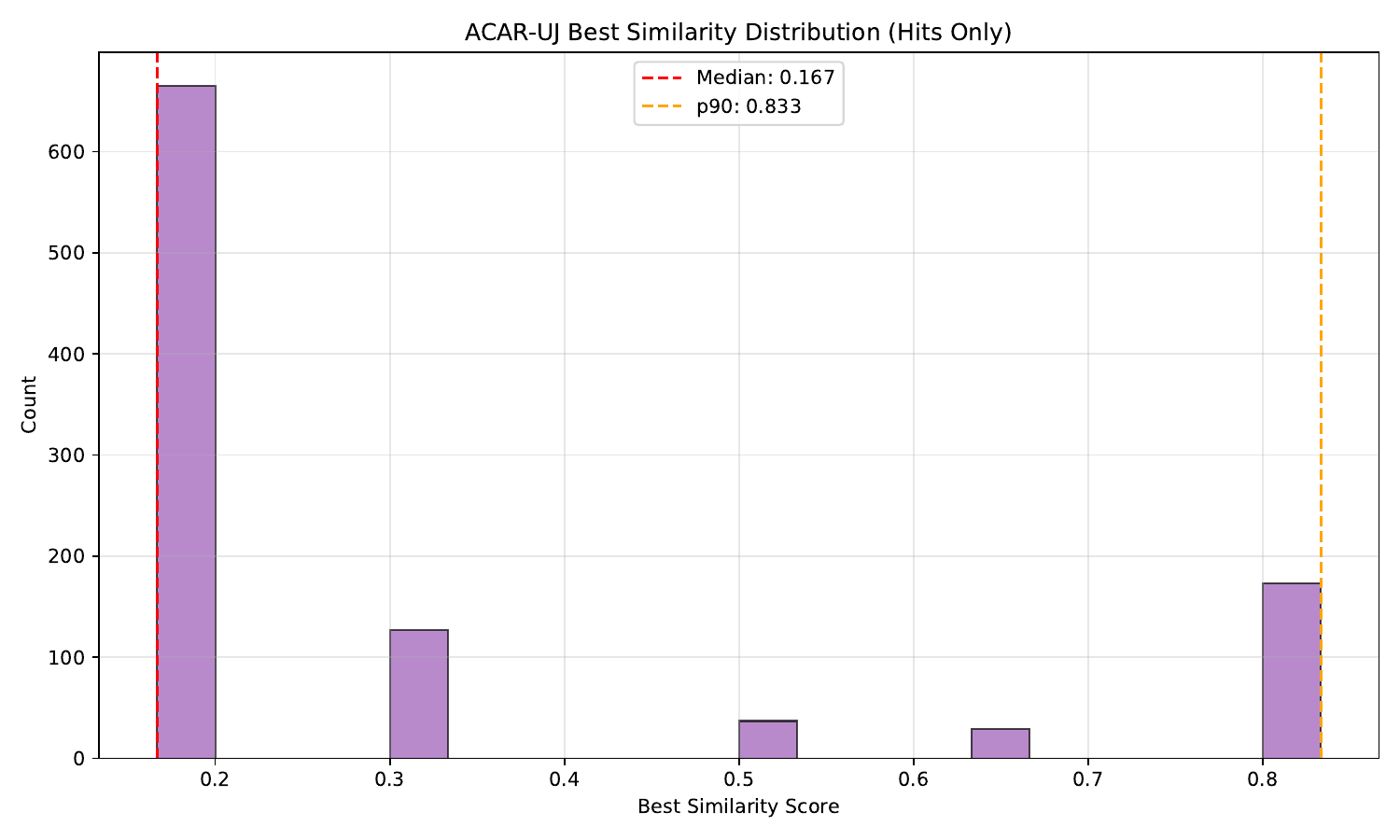}
\caption{\textbf{Similarity distribution for retrieved experiences.} Median similarity of 0.167 explains why retrieval hurt.}
\label{fig:similarity}
\end{figure}

\acar{}-UJ \emph{decreased} accuracy by 3.4pp overall (Table~\ref{tab:uj}). The experience store (837 entries) achieved high hit rates (84-100\%) but \textbf{median similarity was only 0.167} (Figure~\ref{fig:similarity}). Most retrieved experiences were weakly relevant, injecting noise rather than grounding.

\begin{figure}[t]
\centering
\includegraphics[width=0.9\columnwidth]{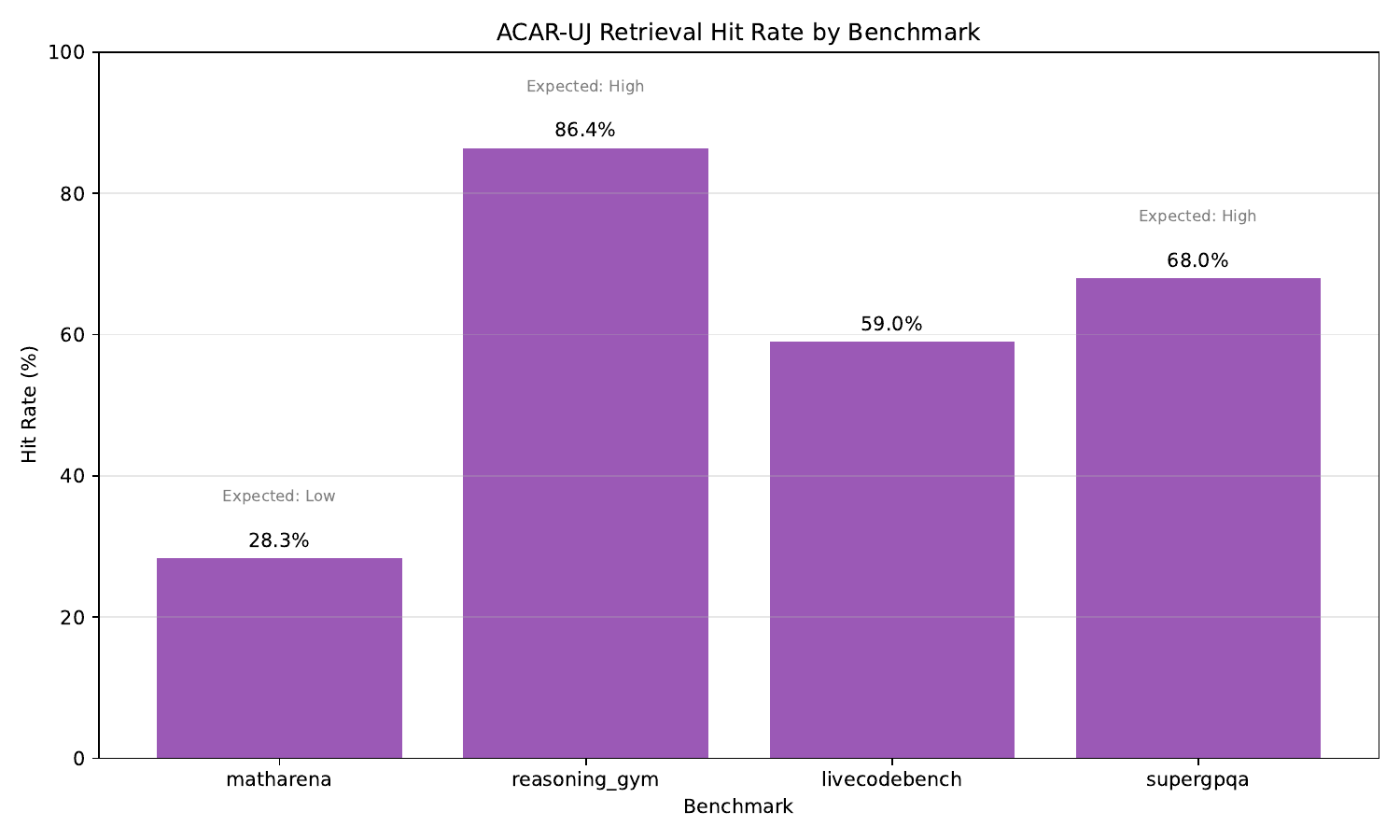}
\caption{\textbf{Retrieval hit rate by benchmark.} High hit rates did not predict retrieval utility.}
\label{fig:hitrate}
\end{figure}

\textbf{Implication}: Retrieval augmentation requires task-aligned stores with similarity thresholds $>$0.7.

\subsection{Agreement-But-Wrong is Unrecoverable}

When all probe samples agree on an incorrect answer ($\sigma=0$), \acar{} routes to single-agent mode and cannot recover. This ``agreement-but-wrong'' failure is intrinsic to self-consistency: if models consistently produce the same wrong answer, no amount of ensembling helps. The 8pp gap between \acar{}-U and Arena-3 quantifies this irreducible ceiling.

\subsection{Attribution Proxies Do Not Work}

We attempted to estimate model contribution using proxy signals (response similarity to final answer, entropy of outputs, agreement patterns). These showed weak correlation with ground-truth leave-one-out values computed via counterfactual runs. Practical attribution requires explicit counterfactual computation.

%==============================================================================
% 7. DISCUSSION
%==============================================================================
\section{Discussion}

\paragraph{What worked.}
(1) $\sigma$-based routing is robust without learned components.
(2) Selective escalation provides 70\% cost reduction on easy tasks.
(3) Infrastructure discipline enables reproducible comparison.

\paragraph{What failed.}
(1) Attribution proxies did not correlate with ground truth.
(2) Naive retrieval hurt; ``more context'' is not automatically beneficial.
(3) Agreement-but-wrong bounds achievable accuracy.

\paragraph{Lessons learned.}
Three findings generalize beyond \acar{}:

\textbf{Self-consistency routing has fundamental limits.} When models unanimously agree incorrectly, no downstream ensemble can recover. Any system relying on agreement-based difficulty estimation will exhibit this ceiling.

\textbf{Retrieval requires semantic alignment.} High hit rate does not imply utility. Without similarity thresholds, retrieval injects noise.

\textbf{Attribution requires counterfactuals.} Post-hoc attribution from observational data does not work. Practical attribution needs explicit counterfactual runs or architectures that make contribution explicit by design.

%==============================================================================
% 8. LIMITATIONS
%==============================================================================
\section{Limitations}

\begin{itemize}[leftmargin=*, itemsep=2pt]
    \item \textbf{Model set}: Three models from major providers; may not generalize to open-source models.
    \item \textbf{Benchmark bias}: SuperGPQA dominates (66\% of tasks).
    \item \textbf{No learned routing}: Learned routers may outperform on specific distributions. We consider this a feature for measurement validity.
    \item \textbf{Code equivalence}: LiveCodeBench escalation is inflated by syntactically different but semantically equivalent outputs.
\end{itemize}

%==============================================================================
% 9. CONCLUSION
%==============================================================================
\section{Conclusion}

We presented \acar{}, adaptive routing using self-consistency variance. The primary contribution is a \emph{measurement methodology}: $\sigma$-based routing exceeds two-model baselines on 54\% fewer full-ensemble calls, while naive retrieval \emph{hurts} without task-aligned stores. We documented three assumptions that fail in practice: self-consistency cannot recover from unanimous incorrect agreement; retrieval with high hit rate does not improve performance; attribution cannot be estimated from proxy signals. With 7,550+ auditable runs and regenerable figures, we provide falsifiable baselines for future research.

%==============================================================================
% AI ASSISTANCE DISCLOSURE
%==============================================================================
\paragraph{AI Assistance Disclosure.}
Large language models were used as development and editorial aids in this work. Claude was used to assist in the development of the experimental test harness and in writing and debugging unit tests and evaluation code. ChatGPT was used for drafting support, structural editing, and clarity improvements in the manuscript text.

All experimental design decisions, benchmarks, hypotheses, implementations, executions, result validation, interpretations, and conclusions were conceived, executed, and verified by the author. The author takes full responsibility for the correctness and integrity of the work.

%==============================================================================
% REFERENCES
%==============================================================================
\bibliographystyle{plainnat}

%==============================================================================
% APPENDIX
%==============================================================================
\appendix

\section{Reproducibility}

All experimental artifacts are structured for independent verification:

\begin{itemize}[leftmargin=*]
    \item All runs logged with seed, prompt hash, environment fingerprint
    \item Complete \texttt{runs.jsonl} with per-task decision traces (7,550+ runs)
    \item Audit reports: zero parse errors across all runs
    \item 208 unit tests for infrastructure validation
    \item All figures regenerable from released artifacts
\end{itemize}

All code, execution artifacts, and figure regeneration scripts are publicly available at:
\url{https://github.com/mechramc/ACAR-TeamLLM}

\section{Artifact Manifest}

\begin{verbatim}
artifacts/phase22_acar_u/
  runs.jsonl          # 1,510 ACAR-U runs
  figures/            # All main result figures

artifacts/phase22_acar_uj/
  runs.jsonl          # 1,510 ACAR-UJ runs
  figures/            # Retrieval analysis figures

datasets/
  arena_3model_runs.jsonl    # Arena-3 baseline
  arena_2model_runs.jsonl    # Arena-2 baseline
  single_model_runs.jsonl    # Single-model baseline
\end{verbatim}

\end{document}